\documentclass{article}

\usepackage{subfigure}

\usepackage{spconf,amsmath,epsfig}
\usepackage{graphicx}
\usepackage{times}
\usepackage{epsfig}
\usepackage{amsmath}
\usepackage{amssymb}
\usepackage{booktabs}
\usepackage{multirow}
\usepackage{verbatim}
\usepackage{appendix}
\usepackage{adjustbox}
\usepackage{pifont}
\usepackage{comment}
\usepackage{color}
\usepackage{bm}

\newcommand\blfootnote[1]{%
  \begingroup
  \renewcommand\thefootnote{}\footnote{#1}%
  \addtocounter{footnote}{-1}%
  \endgroup
}

\renewcommand\paragraph[1]{\noindent{\textbf{#1}}~~~}

\let\OLDthebibliography\thebibliography
\renewcommand\thebibliography[1]{
  \OLDthebibliography{#1}
  \setlength{\parskip}{0pt}
  \setlength{\itemsep}{0pt plus 0.3ex}
}

\usepackage[pagebackref,breaklinks,colorlinks]{hyperref}

\begin{document}\sloppy

\def\x{{\mathbf x}}
\def\L{{\cal L}}

\title{Self-supervised Point Cloud Completion on Real Traffic Scenes via Scene-concerned Bottom-up Mechanism}
%
\name{Yiming Ren$^{1}$, Peishan Cong$^{1}$,Xinge Zhu$^{2}$,Yuexin Ma$^{1,3\dagger}$}
\address{$^{1}$ShanghaiTech University $^{2}$Chinese University of Hong Kong \\ $^{3}$Shanghai Engineering Research Center of Intelligent Vision and Imaging\\
{\{renym1,congpsh,mayuexin\}@shanghaitech.edu.cn}
}

\maketitle
\section{Abstract}
\label{sec:abstract}
Real scans always miss partial geometries of objects due to the self-occlusions, external-occlusions, and limited sensor resolutions. Point cloud completion aims to refer the complete shapes for incomplete 3D scans of objects. Current deep learning-based approaches rely on large-scale complete shapes in the training process, which are usually obtained from synthetic datasets. It is not applicable for real-world scans due to the domain gap. In this paper, we propose a self-supervised point cloud completion method (TraPCC) for vehicles in real traffic scenes without any complete data. Based on the symmetry and similarity of vehicles, we make use of consecutive point cloud frames to construct vehicle memory bank as reference. We design a bottom-up mechanism to focus on both local geometry details and global shape features of inputs. In addition, we design a scene-graph in the network to pay attention to the missing parts by the aid of neighboring vehicles. Experiments show that TraPCC achieve good performance for real-scan completion on KITTI and nuScenes traffic datasets even without any complete data in training. We also show a downstream application of 3D detection, which benefits from our completion approach.

\vspace{-2ex}

\blfootnote{$\dagger$: Corresponding author.}

\section{Introduction}
\label{sec:introduction}
Using scan sensors, such as RGBD cameras or LiDARs, to capture the point cloud of objects or scenes becomes an important mean to obtain the real-world information~\cite{li2019aads,Dai2017ScanNetR3}. However, in real applications, like autonomous driving, it is unpractical to scan the traffic scenes from different views to acquire complete shapes of vehicles. Almost all objects have incomplete point clouds due to the self-occlusion, external-occlusion, and limited sensor resolution. Complementing the missing geometries of point cloud is significant for many downstream tasks~\cite{cong2021input,zhu2021cylindrical}, including scene understanding, navigation, etc. 

Previous approaches~\cite{zhao20193d,wang2020cascaded,zhao20193d,wen2020point,zhang2020detail,xie2020grnet} of point cloud completion are in supervised manner, which heavily rely on ground truth for training the neural network. They utilize synthetic CAD models from ShapeNet~\cite{Chang2015ShapeNetAI} to create a large-scale dataset containing pairs of partial and complete point clouds. However, for real scans, it is difficult to access high-quality paired training data. Some recent works~\cite{chen2019unpaired,wu2020multimodal,zhang2021unsupervised,wen2021cycle4completion} tend to learn a completion network by an unpaired way, which focus on establishing the shape correspondence in latent space between the incomplete point clouds and complete ones. However, such methods still need abundant synthetic complete data in the training process. The gap between synthetic data and real scans and limited training samples of objects make such methods difficult to be extended to real applications. For autonomous driving, LiDAR is the widely-used sensor to capture point clouds of traffic scenes. Due to the working principles of LiDAR, the point cloud has its own characteristics, like the ring sweeps, the varying sparsity, etc. It is more challenging to use synthetic data to solve the LiDAR point cloud completion. Diverse vehicle shapes in the real world is also a thorny problem for above methods. 

\begin{figure}[t]
\centering
\includegraphics[width=1\columnwidth]{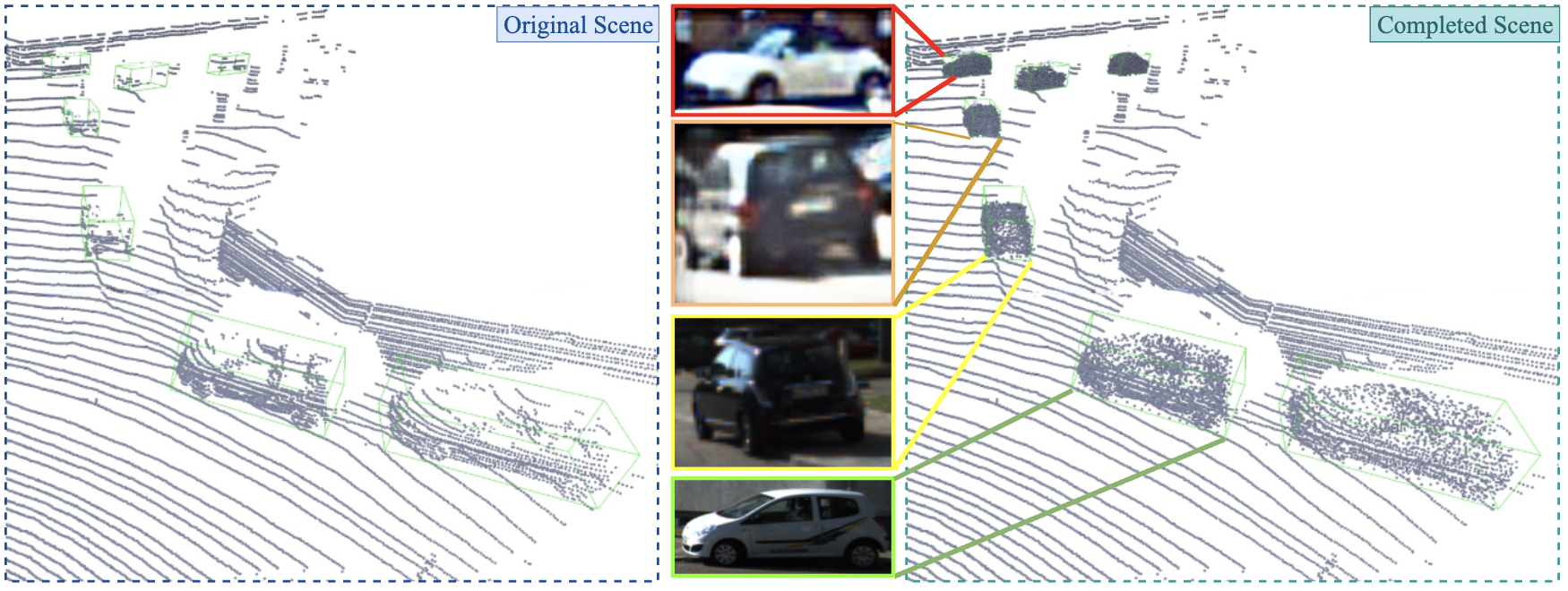}
	\caption{(Left) Raw LiDAR scan from KITTI and (Right) completion results by TraPCC. We attach several images of corresponding cars for reference. It shows that our method can achieve reasonable predictions even for extremely sparse partial vehicle point cloud and approach referred ground truth shown in the images.}
	\label{fig:tensor}
\vspace{-2ex}
\end{figure}

Actually, without seeing one object in 360 degree, human can still imagine the whole appearance of the object based on the partial information and the memory of similar objects. For vehicles in captured traffic scenes, we only have their partial point cloud from one view. But we have large-scale partial point clouds of other vehicles from other views. In addition, we capture the traffic scene consecutively, the temporal information is also helpful for strengthening the memory. Motivated by these findings, we propose a self-supervised approach (TraPCC) for point cloud completion on real traffic scans via utilizing the similarity and symmetry of vehicles. To focus on the local geometries and global shape and size features of the input partial point cloud, we design a bottom-up mechanism by forecasting dense point clouds for the partial bodies at first and then predicting the complete point cloud based on these results provided by the first stage. In one scene, neighboring vehicles can guidance the prediction from two aspects, where one is that they locates in the same view of LiDAR, so they have similar missing parts due to the self-occlusion; the other is that they may occlude for each other. Considering that these hints can help the network pay attention to the missing regions of the point clouds, we propose a scene graph to mine this characteristic, thus improving the capacity of TraPCC. We evaluate our method on KITTI~\cite{geiger2013vision} tracking dataset and nuScenes~\cite{caesar2020nuscenes} dataset. As Figure.~\ref{fig:tensor} shows, our method can achieve reasonable completion results for all the vehicles detected in the scene. Because there is no paired ground truth for strict quantitative analysis, we use alternative metrics~\cite{yuan2018pcn} to illustrate the fidelity and the effectiveness of our method. Ablation study denotes the effectiveness of each module of our network. Experiments also show that TraPCC can act as a filter and benefit the downstream 3D detection task. Our contribution can be concluded as follows.

\begin{enumerate}
\item[1)] We propose a self-supervised point cloud completion for vehicles in traffic scenes without any complete vehicle models, which is applicable for real scans.
\vspace{-1.5ex}
\item[2)] We design a bottom-up mechanism to focus on the missing parts from local and global aspects to boost the completion performance.
\vspace{-1.5ex}
\item[3)] We propose scene graph in our network to make use of the guidance from neighboring vehicles.
\vspace{-1.5ex}
\item[4)] Our method achieve good performance on real scan datasets and can benefit 3D detection tasks for autonomous driving.
\end{enumerate}

\section{Related Work}
\label{sec:related work}
3D shape completion has aroused a wide concern~\cite{wang2020cascaded,xie2020grnet} in recent years. We first review related traditional methods, and then introduce deep learning-based methods, which is categorized into supervised and unpaired approaches.

\paragraph{Traditional Methods} Early works can be divided into two categories, including geometry-based methods and alignment-based methods. The former~\cite{nguyen2016field,zhao2007robust,berger2014state,sung2015data}, utilizes geometry cues of the partial input to fill holes in the incomplete scans, and the latter~\cite{nan2012search,shen2012structure,li2010analysis} usually retrieves complete shapes from a large database by optimization. These traditional methods requires expensive computation during inference and are sensitive to noise, which is not applicable for the incomplete data captured from the real world.

\paragraph{Supervised Methods} Deep learning-based methods~\cite{zhao20193d} take advantage of the neural networks to extract features from the incomplete input shape and predict complete representation. Point cloud-based shape completion~\cite{yuan2018pcn} has rapid development compared with meshes and voxel grids. Many methods \cite{wang2020cascaded,zhao20193d,wen2020point,zhang2020detail,xie2020grnet,xu2019depth} achieve impressive predicted results. To preserve the spatial arrangements of the input, PF-Net~\cite{huang2020pf} only predicts the missing part of the point cloud rather than the whole object. SK-PCN~\cite{nie2020skeleton} maps the partial scan to complete surface with the aid of meso-skeleton. RL-GAN-Net~\cite{sarmad2019rl} proposes a reinforcement learning agent controlled
GAN for completion and PoinTr~\cite{yu2021pointr} adopts a transformer encoder-decoder
architecture. These works, however, are all in supervised manner, which requires incomplete and complete paired data for training deep neural networks. Actually, it is not easy to access such high-quality paired data in real-world applications. Such methods perform well for synthetic data but not for real scans without paired ground truth.

\paragraph{Unpaired Methods} To reduce the gap between synthetically-generated data and real-world data, some methods loose the paired data constraint in supervised methods to unpaired data training. ~\cite{stutz2018learning} creates a latent space created for clean and complete data and uses an unsupervised maximum likelihood loss for the measurement. Pcl2pcl~\cite{chen2019unpaired} trains two separate auto-encoders for constructing partial and complete shapes respectively, and uses GAN setup to learn a mapping between these two latent spaces. The follow-up work~\cite{wu2020multimodal} extends it to multiple predictions. ShapeInversion~\cite{zhang2021unsupervised} applies GAN inversion to shape completion and achieve more faithful results. Cycle4Completion~\cite{wen2021cycle4completion} designs two cycle transformations to establish the geometric correspondence
between incomplete and complete shapes from both directions and is superior in unpaired completion methods. Even though these methods do not reply on paired data any more, they still use large-scale synthetic complete 3D shapes belonging to the same kind in the training procedure. Without any complete 3D models, our unsupervised method only takes advantage of the real scans for point cloud completion, which is practical for real applications and has better generalization capability.

\section{Methodology}
\label{sec:method}
\begin{figure*}[ht!]
\centering
\includegraphics[width=2.1\columnwidth]{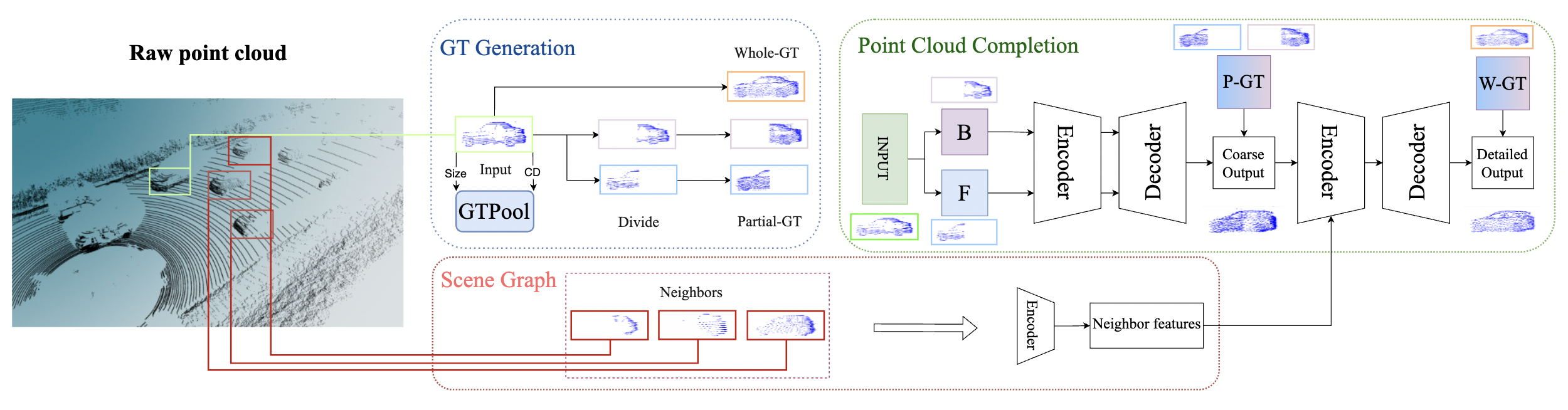}
	\caption{The pipleine of TraPCC. The whole framework consists two main modules: Whole-GT and Partial-GT generation and point cloud completion network. We first divide the input point cloud into B(back) and F(front) two parts, then take Partial-GT and Whole-GT as the supervision information. Neighbor features are considered by scene graph to improve the final prediction.}
	\label{fig:framework}
\end{figure*}
\subsection{Overview}

The framework overview is illustrated in Figure.~\ref{fig:framework}. The whole framework consists of two main modules:
1) Ground truth generation. For real scans, we cannot get real complete point cloud as ground truth in training. Based on the similarity and symmetry properties of vehicles, we construct a memory pool to store seen vehicles in raw point cloud as our pseudo-ground-truth database, which will be called ground truth (GT) in the following. 2) TraPCC network. It is in a bottom-up mechanism, which can focus on the local geometries and global shape and size features of the input partial point cloud. Scene graph is designed for enhancing the completion performance by considering the guidance from neighboring objects.


\begin{figure}[ht]
\centering
\includegraphics[width=1\columnwidth]{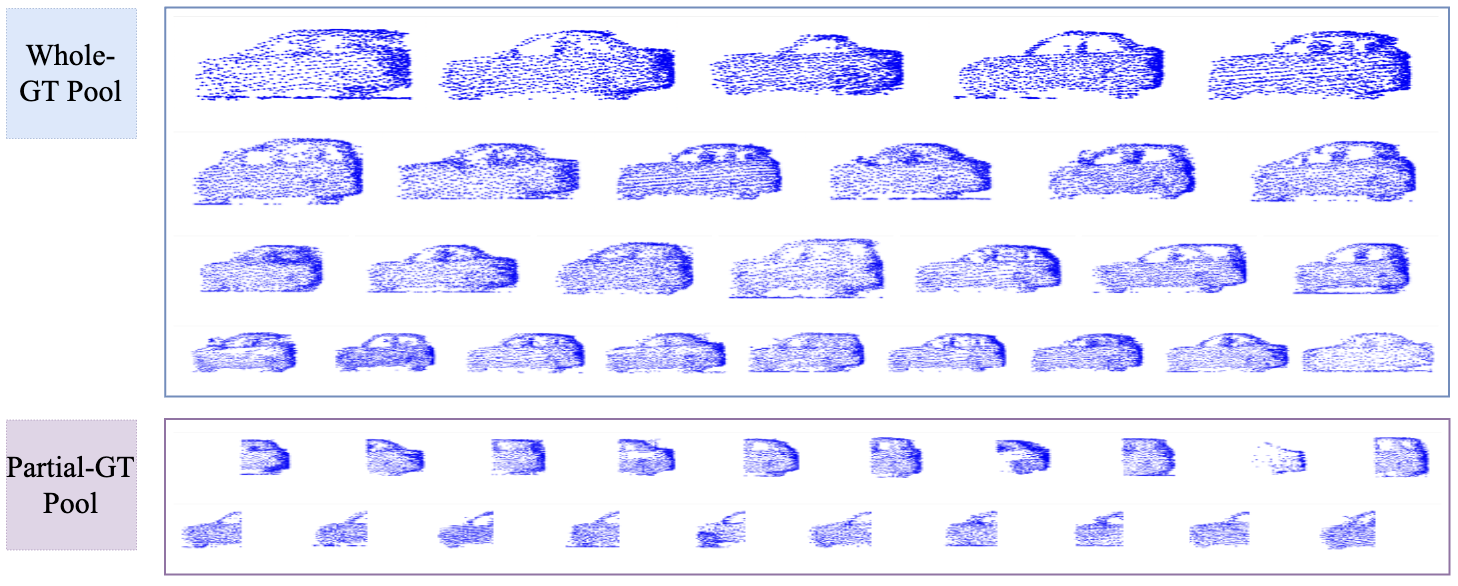}
	\caption{Ground Truth Pool in KITTI tracking dataset. It contains relative complete real scans of vehicles with different shapes.}
	\label{fig:GTPool}
\vspace{-3ex}
\end{figure}
\subsection{Ground Truth Generation}




\paragraph{Ground Truth Pool} We extract vehicles from point cloud scenes according to 3D bounding boxes obtained by 3D detection. For instances occur in sequential frames, they may be captured from multiple views. We combine the point cloud in consecutive frames according to the tracking ID and motion compensation and gain a relative complete point cloud. We take the data with more than 1024 points as GT candidates. Then we use the 2D contours of the three views (side, front, and bird-eye view) of the instance to decide whether it is relative complete. We map the point cloud to the 2D plane from one view and sample evenly along the coordinate axis of the object coordinate system. We check the number of points in four quadrants to judge whether the shape is evenly distributed and then select objects with relatively evenly distributed points to the Whole-GT pool according to the threshold by statistics. We also construct a Partial-GT pool to store the heads and tails of vehicles in the same way. The visualization of the GT-Pool is shown in Figure.~\ref{fig:GTPool}. 



\paragraph{Ground Truth Generation} The main steps to generate GT is illustrated in Figure~\ref{fig:framework}. For objects extracted from raw point cloud, we take the similarity of size and shape into account to search suitable complete point cloud in GT-Pool. The shape similarity is calculated by Chamfer Distance (CD)~\cite{fan2017point} of the input point cloud and GT. The function of CD is 
\begin{equation}
\begin{aligned}
d_{CD} (S_{1},S_{2})=\frac{1}{|S_{1}|}\sum_{x \in S_{1}}\min_{y \in S_{2}}\vert x-y \vert_{2}^2 \\
+\frac{1}{|S_{2}|}\sum_{y \in S_{2}}\min_{x \in S_{1}}\vert y-x \vert_{2}^2, 
\end{aligned}
\label{equ:cdloss}
\end{equation}
where $S_{1}$ and $S_{2}$ denotes input point cloud and GT respectively, $x$ and $y$ represents the 3D coordinates of points. For extremely sparse input, we modify the CD-Loss to better calculate the similarity from sparse point cloud to dense point cloud. The modified CD-Loss (defined as MCD loss) is shown as 
\begin{equation}
\begin{aligned}
d_{MCD} (S_{1},S_{2})=\min(d_{CD} (S_{1},S_{2}))\\
=\frac{1}{S_{1}}\sum_{x \in S_{1}}\min_{y \in S_{2}}\vert x-y \vert_{2}^2.
\end{aligned}
\label{equ:mcdloss}
\end{equation}

\subsection{Point Cloud Completion}
We introduce detailed TraPCC network in this section. The main architecture is a bottom-up point cloud completion, which take advantages of fine-grained and global features to guide the completion of missing parts. Moreover, we design scene graph in TraPCC to make use of the scene information from neighbours to consider the effect of self-occlusion and external-occlusion.


To take advantage of partial and complete ground truth to extract local fine-grained feature and global shape information respectively, we design a two-stage completion network to get the coarse results and detailed results in order.
We use a modified PCN~\cite{yuan2018pcn} as backbone, which uses two PointNet modules in encoder and we discard the coarse-to-fine decoder for the reason that the ground truth generated in original LiDAR sparse point cloud are not so dense as synthetic models.

In the first stage (P-Net), incomplete point cloud is divided into front and back parts according to the bounding box labeled in detection task with the supervision of Parial-GT. And then, these two parts are stitched together as a whole coarse object. If there is no points in the front or back parts, there will not be a corresponding partial prediction result. The second stage (C-Net) is designed to refine the coarse object obtained from the last stage in order to take global shape features into consideration with the supervision of Whole-GT.


The emitted light from LiDAR will bounce back when hitting an obstacle. According to LiDAR working principle, 3D objects have incomplete point cloud due to self-occlusion caused by itself and external-occlusion caused by surrounding objects (especially the object between LiDAR and target object). For self-occlusion cases, the instance 
around the target object can offer similar partial spatial information. For external-occlusion cases, the object nearer LiDAR can provide a clue to the occluded part of target object. We obtain guidance through occluded information from neighbours by constructing a scene-graph shown in Figure ~\ref{fig:framework}. 

The scene graph is designed by taking the input data as the center of a circle and extracting the nearest k neighbors (in experiment, we set k=3) of the input object within a certain radius. As a result, TraPCC can effectively optimize the prediction result with the guidance of neighbour features.
 
 

\begin{figure}[t]
\centering
\includegraphics[width=0.95\columnwidth]{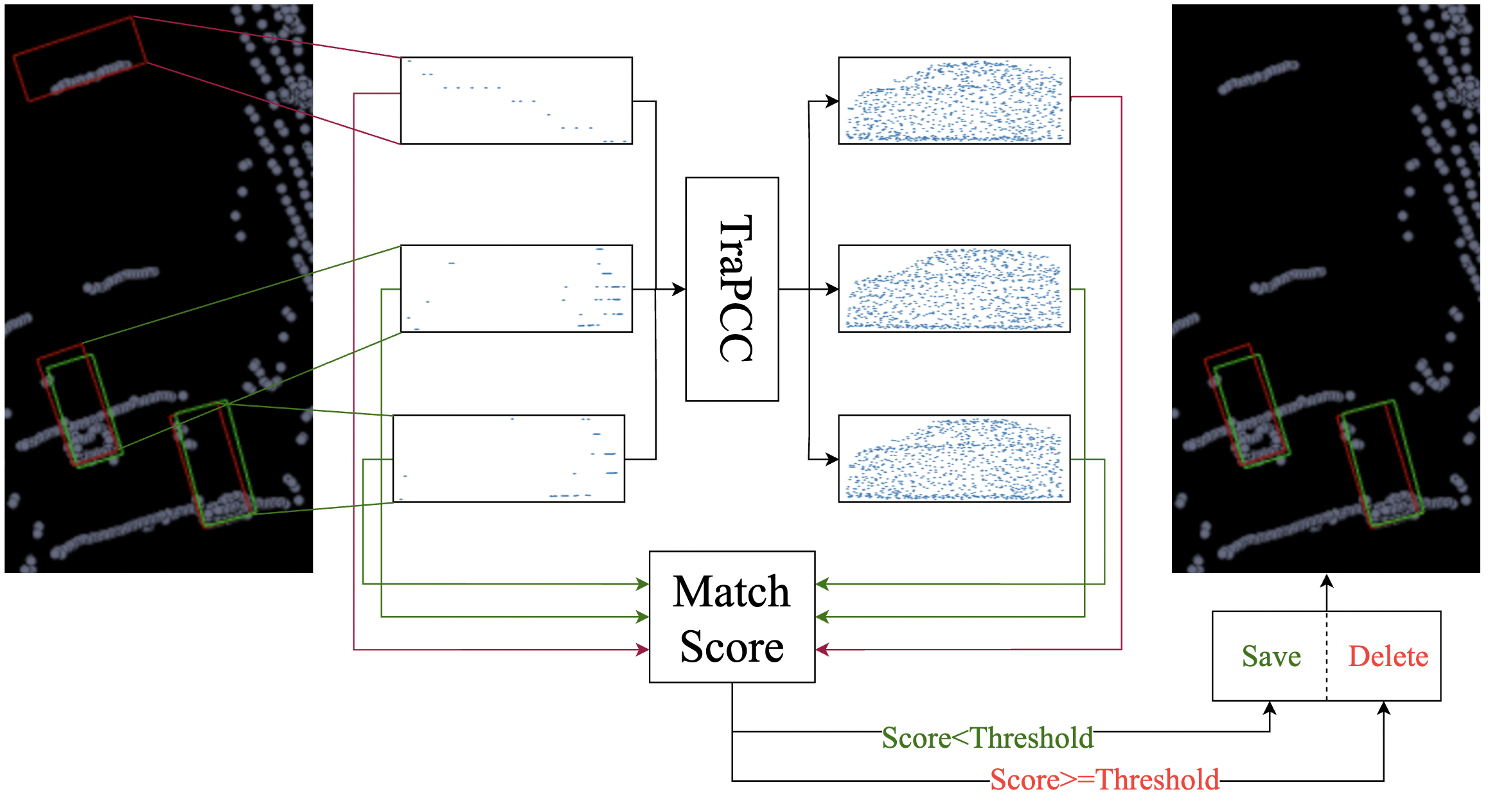}
	\caption{TraPCC for improving detection precision. The whole process consist three steps: (A) Detection: Using detection network to get initial detected vehicles. (B) Point cloud completion: complete the objects extracted from the detection stage and compute the shape difference with the original input object to distinguish whether the point cloud is a vehicle or noise. (C) Filter: Delete the object whose score is higher than the threshold. Green Bounding boxes in the figure are ground truth and red ones are detection results.}
	\label{fig:improve}
\vspace{-2ex}
\end{figure}

\subsection{Loss Function}
Our proposed loss function consists of three terms, the P-Net completion CDloss between coarse partial output $Y_{coarse}$ and partial ground truth $Y_{P-GT}$, the C-Net completion CDloss loss between detailed whole output $Y_{detailed}$ and the whole ground truth $Y_{W-GT}$, and input-output comparison loss (defined as MCD-Loss) between the input data$X_{input}$ and $Y_{detailed}$.
\begin{equation}
\begin{aligned}
L(&Y_{coarse},Y_{detailed},X_{input},Y_{gt})= \\
&d_{CD}(Y_{coarse},Y_{P-GT})+d_{CD}(Y_{detailed},Y_{W-GT})\\
&+d_{MCD}(Y_{detailed},X_{input})
\end{aligned}
\label{equ:loss}
\end{equation}

Considering the fact that object point cloud in the real LiDAR scenes is sparse and may miss large parts of body, MCD-Loss strengthens the connection between the input and the output point clouds in case that the original information will be neglected after being trained on the generated ground truth. It can guarantee that our prediction result is not limited by generated ground truth and keeps the characteristics of the input.
\subsection{Adaptation for Downstream Tasks}
We evaluate the effectiveness of our self-supervised point cloud completion in detection task with the main steps illustrated in Figure.~\ref{fig:improve}. Specifically, we use 3D detection network to extract the objects from raw point cloud and process point cloud completion by TraPCC to get complete vehicles. Then, we compare the completed results with the raw object point cloud by MCD and 2D contour differences from three views. We set a threshold of the score, which describes the shape differences, to filter the detected objects. The detected object with the scores above the threshold are determined as a mismatch. In other words, the shape of the detected object is far from vehicle shapes and is most likely some noise points, which should be deleted from the detection results. Experiments show that we can improve the detection precision by filtering false predictions with TraPCC.

\section{Experiments}
\label{sec:experiments}
In this section, we introduce the implementation details and evaluate the performance of TraPCC. Ablation studies are conducted to validate each component of our approach. We also evaluate the improvement of our TraPCC in detection tasks.
\subsection{Dataset}

 We use two datasets in our experiments, KITTI\cite{geiger2013vision} tracking dataset and nuScenes\cite{caesar2020nuscenes} dataset, both of which provide the ID information that can be used to generate the GT-Pool. KITTI tracking dataset has twenty sequences with different numbers of vehicles, we use one sequence with 5242 vehicles to train the model and one sequence with 1354 vehicles to test the model. For nuScenes dataset, we utilze 3262 objects for training and 707 objects for testing.


\subsection{Implementation Details}

3D object detection uses SECOND\cite{yan2018second} and PointPillar\cite{lang2019pointpillars} as backbones on KITTI tracking dataset and nuScenes. The threshold setting of matching score and detection score is 0.3 and 0.5 respectively in KITTI and 0.4 and 0.5 in nuScenes, meaning we only filter the false prediction by matching score when detection score is below detection threshold.

\noindent\textbf{Evaluation Metric} To evaluate the completion results of our model, we design three kinds of loss. \textbf{L-G} is the CD between the ground truth (partial-GT and complete-GT) and prediction result on average. \textbf{L-I} is MCD, reflecting the fidelity of the result and \textbf{L-S} illustrates the shape difference from the side view, which is computed by the CD of the 2D projected shapes of GT and input. And meanCD is the average of the above three kinds of Chamfer distances.
\begin{table}[t]
    \centering
     \caption{TraPCC completion results on KITTI nad nuScenes.
    }
    \begin{tabular}{c|ccc|c}
         Dataset & L-G & L-I & L-S & meanCD \\ \hline
         nuScenes\cite{caesar2020nuscenes} & 0.1232 & 0.1098 & 0.1192 & 0.1175 \\ \hline
         KITTI\cite{geiger2013vision} & 0.0951 & 0.1031 & 0.0939 & 0.0973 \\\hline
    \end{tabular}
   
    \label{tab:result}
\end{table}

\subsection{Results}

\noindent\textbf{Completion Result on KITTI and nuScene}
Our method totally use the data from the raw scan without any template model or complete synthetic ground truth in the training process, it is unfair to compare our method with paired or unpaired supervised methods. We only show our results in this section. We train and test the model on KITTI tracking dataset and nuScenes dataset, respectively. The result shown in Table.\ref{tab:result} illustrate the small shape difference between the predicted results and GT as well as keeping the fidelity. 

\noindent\textbf{Filter Results for 3D Detection} We report the performance on KITTI validation set in Table.~\ref{tab:detection}.
 The detection precision can be improved especially for hard cases, which illustrates TraPCC can definitely help to remove false detected results.
We also use different training datasets to see the generalization capability of our method. We train the completion network on nuScenes dataset and use it to boost the detection precision on KITTI dataset. Especially for the PointPillar-based detection, the cross-dataset training has obvious improvement.

\begin{table}[ht]\small
	\centering
	\caption{Detection performance on KITTI validation set. K means training on kitti tracking dataset. N means training on nuScenes dataset. We show the results on SECOND and PointPillar two detection backbones.}
	\setlength{\tabcolsep}{1.0mm}
	\begin{tabular}{l|ccc|ccc}
		\toprule
	\multirow{2}{*}{Method}&\multicolumn{3}{|c|}{3D Object Detection$(\%)$}&\multicolumn{3}{|c}{BEV Detection$(\%)$} \\ \cline{2-7}
	&Easy&Moderate&Hard&Easy&Moderate&Hard \\ \hline
		SECOND\cite{yan2018second} & 81.78 & 71.75 & 65.91 & 87.06 & 83.24 & 83.00\\
		TraPCC(K) & \textbf{81.89} & \textbf{71.87} & \textbf{65.98} & \textbf{87.09} & \textbf{83.24} & 83.23 \\
		TraPCC(N) & 81.84 & 71.86 & 65.92 & 87.06 & 83.17 &\textbf{83.26} \\
		\hline\hline
		PointPillar\cite{lang2019pointpillars} & 85.00 & \textbf{75.49} & 69.05 & \textbf{88.54} & \textbf{85.99} & 84.45\\
		TraPCC(K) & \textbf{85.02} & \textbf{75.49} & 73.84 & \textbf{88.54} & \textbf{85.99} & 85.64 \\
		TraPCC(N) & 85.00 & \textbf{75.49} & \textbf{73.85} & \textbf{88.54} & 85.95 & \textbf{85.66} \\
\bottomrule 
	\end{tabular}

\label{tab:detection}
\end{table}

\begin{table}[t]
    \centering
     \caption{Ablation studies for different modules on KITTI.
    }
    \begin{tabular}{c|c|c|c}
         Module & C-Net & CP-Net & TraPCC  \\ \hline
         meanCD & 0.0984 & 0.0975 & 
0.0973\\\hline
    \end{tabular}
   
    \label{tab:Ablation}
\end{table}

\begin{figure}[ht]
\centering
\includegraphics[width=0.95\columnwidth]{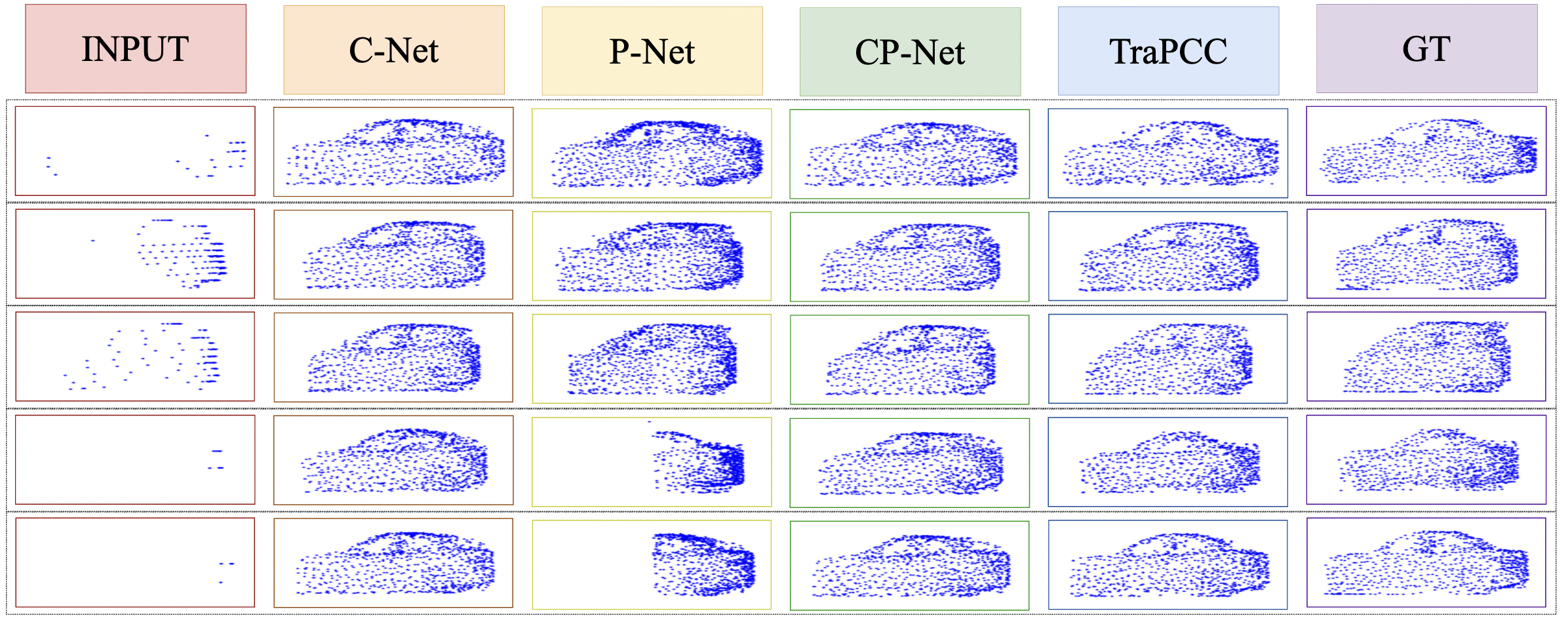}
	\caption{Visualization comparison on KITTI. 
	}
	\label{fig:compare}
\vspace{-2ex}
\end{figure}

\subsection{Ablation Study}
In this section, we perform the ablation study to verify the effect of different components of TraPCC from both quantitative and qualitative aspects as Table.~\ref{tab:Ablation} and Figure.~\ref{fig:compare} shows. The different networks and analysis are as follows.\\


\paragraph{C-Net} We use the PCN~\cite{yuan2018pcn} as the backbone and generate whole complete bodies of input vehicles directly. We can see from the results that it can still get reasonable results by training on our GT-pool.

\paragraph{P-Net} We stitch the front part prediction and back part prediction together to get the results of P-Net. From the visualization, it is easy to see that the local geometries are reconstructed well but the combination result lacks the overall coordination. 

\paragraph{CP-Net} We integrate the P-Net and C-Net to a CP-Net. In other words, CP-Net is the TraPCC without the scene graph (Figure.~\ref{fig:framework}). By considering both the local detailed geometries and global shape features, CP-Net can reduce the meanCD and get better results than P-Net and C-Net.



\paragraph{TraPCC} After adding scene graph, the network can optimize the result by involving more occlusion guidance from neighbor's information. It can be observed that this module contributes an improvement to the network and reach the best performance. From the visualization, it not only approach to the ground truth, but also remove some noise in ground truth.

\begin{figure}[ht]
\centering
\includegraphics[width=0.95\columnwidth]{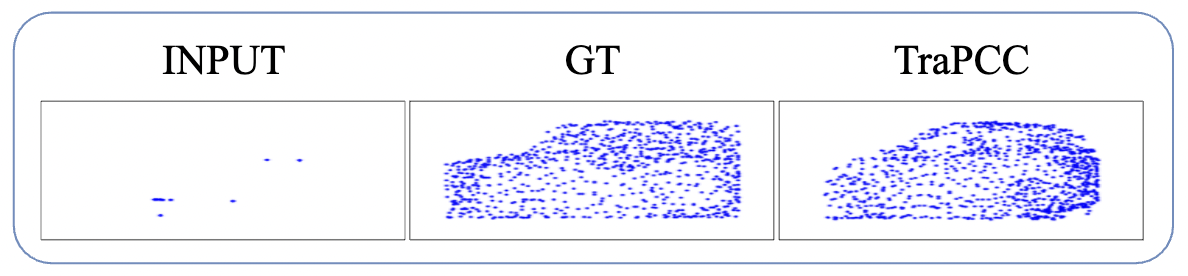}
	\caption{Limitation of TraPCC in very sparse point cloud.
	}
	\label{fig:limitation}
\vspace{-2ex}
\end{figure}
\subsection{Limitations and Future Work}
For LiDAR captured point clouds, apart from the severely occluded situations, the objects far from the LiDAR sensor also have extremely sparse points, which bring difficulties for the completion because of losing information. TraPCC can handle many challenging cases as Figure.~\ref{fig:compare} (last two rows) shows. However it could not work well all the time(Figure.~\ref{fig:limitation}). 
In the future, we aim to refine this work from two aspects. One is using some optimization methods to refine the GT-pool. The other is considering the temporal information in the input to use more geometry features of the target object with more points from consecutive views, which gives more guidance for the completion. 



\section{Conclusion}
We propose a self-supervised method for point cloud completion on real traffic scenes without any template or synthetic data. We select the objects in real scans with enough points and relative complete shapes into our ground truth pool, so that our prediction results approach the real point clouds captured by scan sensors, which solves the gap between synthetic and real data existing in previous methods. We design a bottom-up mechanism to make the network focus on both local and global features. Scene graph is proposed to utilize scene information for completion. Experiments demonstrate the effectiveness and generalization of our method. We also show our method can benefit downstream 3D detection task.

{\footnotesize
\bibliographystyle{IEEEbib}
\bibliography{icme2022template}
}

\end{document}